\pdfoutput=1

\documentclass[11pt]{article}

\usepackage{EMNLP2024}

\usepackage{times}
\usepackage{latexsym}
\usepackage{graphicx} 
\usepackage{amsmath} 
\usepackage{booktabs}
\usepackage[T1]{fontenc}
\usepackage{amsfonts,amssymb}
\usepackage{multicol}
\usepackage{multirow}
\usepackage{fontawesome}
\usepackage{pifont}
\usepackage{bbding}
\usepackage{color, colortbl}
\usepackage{xcolor}
\newcommand{\para}[1]{{\vspace{1.5pt} \bf \noindent #1\hspace{4pt}}}

\usepackage{tcolorbox}
\definecolor{Gray}{gray}{0.9}
\definecolor{demphcolor}{RGB}{144,144,144}
\definecolor{backblue}{RGB}{221,239,251}
\definecolor{backblue1}{RGB}{186,216,242}
\definecolor{backred}{RGB}{244,199,204}
\definecolor{cvprblue}{rgb}{0.21,0.49,0.74}
\usepackage[utf8]{inputenc}

\usepackage{microtype}

\usepackage{inconsolata}

\title{AlignCap: Aligning Speech Emotion Captioning to Human Preferences }

\author{Ziqi Liang$^{1}$\thanks{$^\ast$ Equal contribution.} \ , \ Haoxiang Shi$^{1\ast}$, \ Hanhui Chen$^2$\\
$^1$ University of Science and Technology of China, Hefei, China\\
$^2$ Southern University of Science and Technology, Shenzhen, China\\
\texttt{\{tymlzq, shihaoxiang\}@mail.ustc.edu.cn, hanhuichencn@gmail.com}
}

\begin{document}
\maketitle


\begin{abstract}
Speech Emotion Captioning (SEC) has gradually become an active research task. The emotional content conveyed through human speech are often complex, and classifying them into fixed categories may not be enough to fully capture speech emotions. Describing speech emotions through natural language may be a more effective approach. However, existing SEC methods often produce hallucinations and lose generalization on unseen speech. 
To overcome these problems, we propose \textbf{AlignCap}, which \underline{Align}ing Speech Emotion \underline{Cap}tioning to Human Preferences based on large language model (LLM) with two properties: \textbf{1) Speech-Text Alignment}, which minimizing the divergence between the LLM's response prediction distributions for speech and text inputs using knowledge distillation (KD) Regularization.
\textbf{2) Human Preference Alignment}, where we design Preference Optimization (PO) Regularization to eliminate factuality and faithfulness hallucinations.
We also extract emotional clues as a prompt for enriching fine-grained information under KD-Regularization.
Experiments demonstrate that AlignCap presents stronger performance to other state-of-the-art methods on Zero-shot SEC task.
\end{abstract}

\section{Introduction}
The identification and description of speech emotions play a crucial role in improving communication efficiency. It also aids in understanding the speaker's intentions.
Previous work usually treats emotion acquisition as a classification task, such as Speech Emotion Recognition (SER)~\cite{Emo-DNA, DWFormer,epc_interspeech}, which assigns speech to different emotion categories based on the emotions such as sadness, anger, and happiness contained within the speech. 
However, there may be a mixture of emotions within one utterance, and classifying speech into a single emotion category is not enough to capture the true emotion. Moreover, different annotators may assign different emotion category labels to a piece of speech, leading to the label ambiguity problem in SER task~\cite{SERlabel1,SERlabel2}. This can result in inaccurate emotion labels in existing SER datasets.

\begin{figure}[t]
    \centering
    \includegraphics[width=7.7cm]{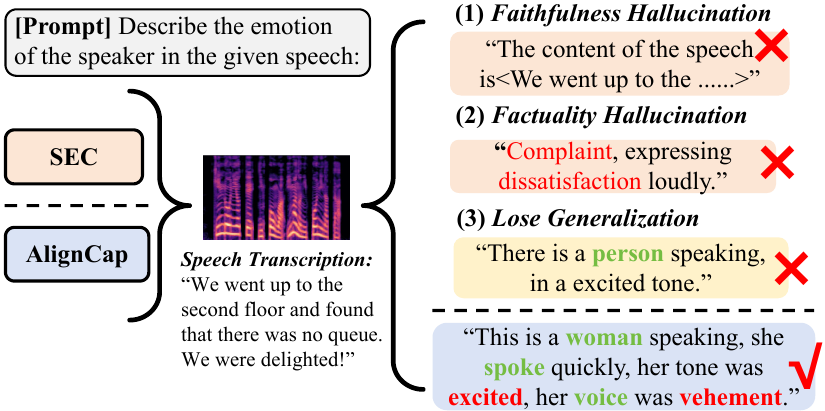}
    \caption{Hallucination and lack of generalization.}
    \vspace{-0.5cm}
    \label{fig1}
\end{figure}

Given the limitations of speech emotion classification, employing natural language descriptions instead of emotion category labels is a more accurately approach.
SECap~\cite{SECap} first proposes a speech emotion captioning framework to describe speech emotions using natural language effectively. It utilizes HuBERT~\cite{HuBERT} as an audio encoder to extract speech features while leveraging mutual information learning to decouple content and emotion-related features. \cite{zeroshotaudiocap2} employs GPT-2 \cite{GPT2} as the decoder to generate captions based on the pre-trained CLAP \cite{clap} audio encoder. \cite{zeroshotaudiocap} exploits OPT \cite{OPT} as the LLM to produce captions that describe the audio content.
However, facing with unseen speech, these methods tend to produce hallucinations of factuality and faithfulness, resulting in false emotional descriptions and responses that are inconsistent with user instructions. 
In addition, the paradigm of text-only training and zero-shot inference on speech like \cite{DCASEWorkshop2023} will lead to training-inference mismatch, resulting in poor model generalization.

In this paper, we propose a novel SEC framework \textbf{AlignCap}, which aims to generate rich and coherent captions while maintaining high consistency with speech emotion.
We design KD-Regularization, which can minimize the distribution gap between LLM's response to speech input and those to corresponding text inputs. It bridges the training-inference mismatch in zero-shot SEC and model generalization is improved. AlignCap is the first to align SEC models to human preferences via PO-Regularization, which eliminates factuality and faithfulness hallucinations of SEC models on unseen speech. We also utilize a acoustic prompt generated from emotional clues to enrich fine-grained information.
To summarize, our contributions are as follows:
\begin{enumerate}
\item We propose KD-Regularization to achieve Speech-Text Alignment and use the KL-divergence of next-token prediction distributions as a measure of alignment. 
\item We propose PO-Regularization to achieve Human Preference Alignment, which generates rich, consistency and rationality emotion descriptions.
\item We analyze the issue of distribution gaps in SEC task and explore various alignment methods to bridge the gap. Experiments demonstrate AlignCap's superiority in both zero-shot and cross-domain scenarios.
\end{enumerate}

\section{Background and Discussion}
The section describes the speech-text distribution gap of traditional SEC methods. To explore this gap, we conducted preliminary experiments to analyze its potential impact on train-inference mismatch and performance degradation. Furthermore, we discuss the impact of modal alignment position on downstream SEC performance.

\begin{figure}[t]
    \centering
    \includegraphics[width=7.4cm]{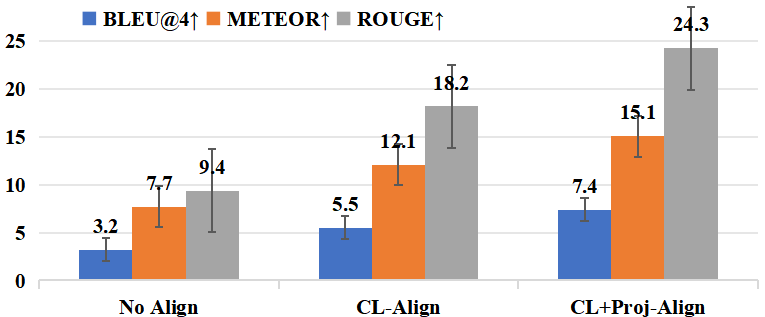}
    \vspace{-0.1cm}
    \caption{Results of different alignment methods.} 
    \vspace{-0.3cm}
    \label{degrade}
\end{figure}

\para{Distribution Gap and Alignment.}
As the creation of speech-caption pairs is costly, traditional SEC methods usually are trained using only text, and employed zero-shot inference on speech. The distribution of speech and text embeddings do not exactly coincide, which degrades the SEC's performance.
To analyze the effect of modal alignment on eliminating distribution gap, we adopt No-alignment, Contrastive Learning alignment (CL-Align) \cite{zeroshotaudiocap2}, and Projection-based alignment (CL+Proj-Align) \cite{zeroshotaudiocap2}, and evaluate the performance of SEC on BLEU@4~\cite{BLEU}, METEOR~\cite{METEOR}, and ROUGE~\cite{ROUGE} metrics.
We conducted three experiments: \textbf{1) No-Align:} Speech encoder of Pre-trained CLAP model \cite{clap} is used to zero-shot inference directly. \textbf{2) CL-Align:} We fine-tuning the text encoder and speech encoder of CLAP using contrastive learning on speech-caption pairs $\mathcal{D}_{s}\!=\!\{(x_{n},y_{n})\}$ which are randomly selected from SEC datasets. \textbf{3) CL+Proj-Align:} Based on CL-Align, we add Projection-based decoding to project the speech embedding into the text embedding space through cosine similarity.
As shown in Fig~\ref{degrade}, captions generated from model with Alignment exhibit superior similarity compared to that No-alignment model. This findings proves that the distribution gap adversely affects the SEC's performance. 

\begin{figure}[b]
    \centering
    \vspace{-0.2cm}
    \includegraphics[width=7.6cm]{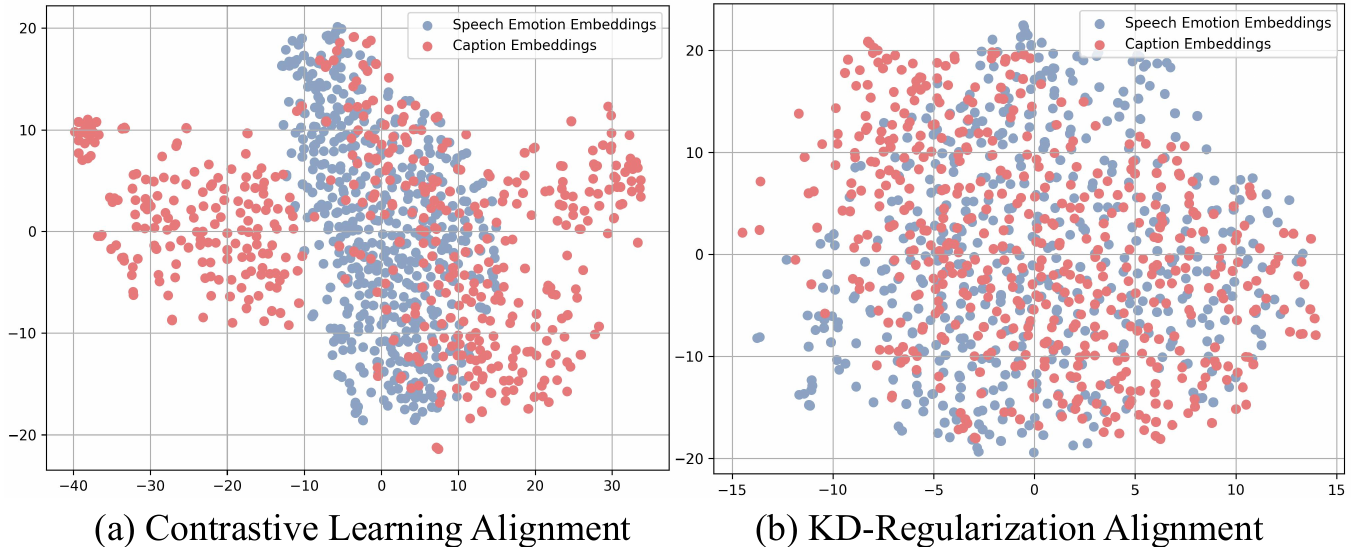}
    \caption{T-SNE visualizations of LLM's output from speech and text input. (a) Align before LLM Decoding. (b) Align after LLM Decoding.} 
    \label{distribution}
\end{figure}

\para{Align before or after LLM Decoding?}
According to \cite{Jiang_2023_CVPR}, complete alignment between modalities is often not the optimal solution for downstream tasks. Such alignment may result in information loss, especially when the information provided by the two modalities differs significantly. 
Traditional SEC models achieve Speech-Text Alignment via fine-tuning encoder on speech-caption pairs, which bridges the distribution gap before LLM decoding. However, complete alignment of speech and text embedding may result in information loss, and it lacks a direct measure for assessing speech-text alignment quality. 

\begin{figure*}[t]
    \centering
    \includegraphics[width=14.0cm]{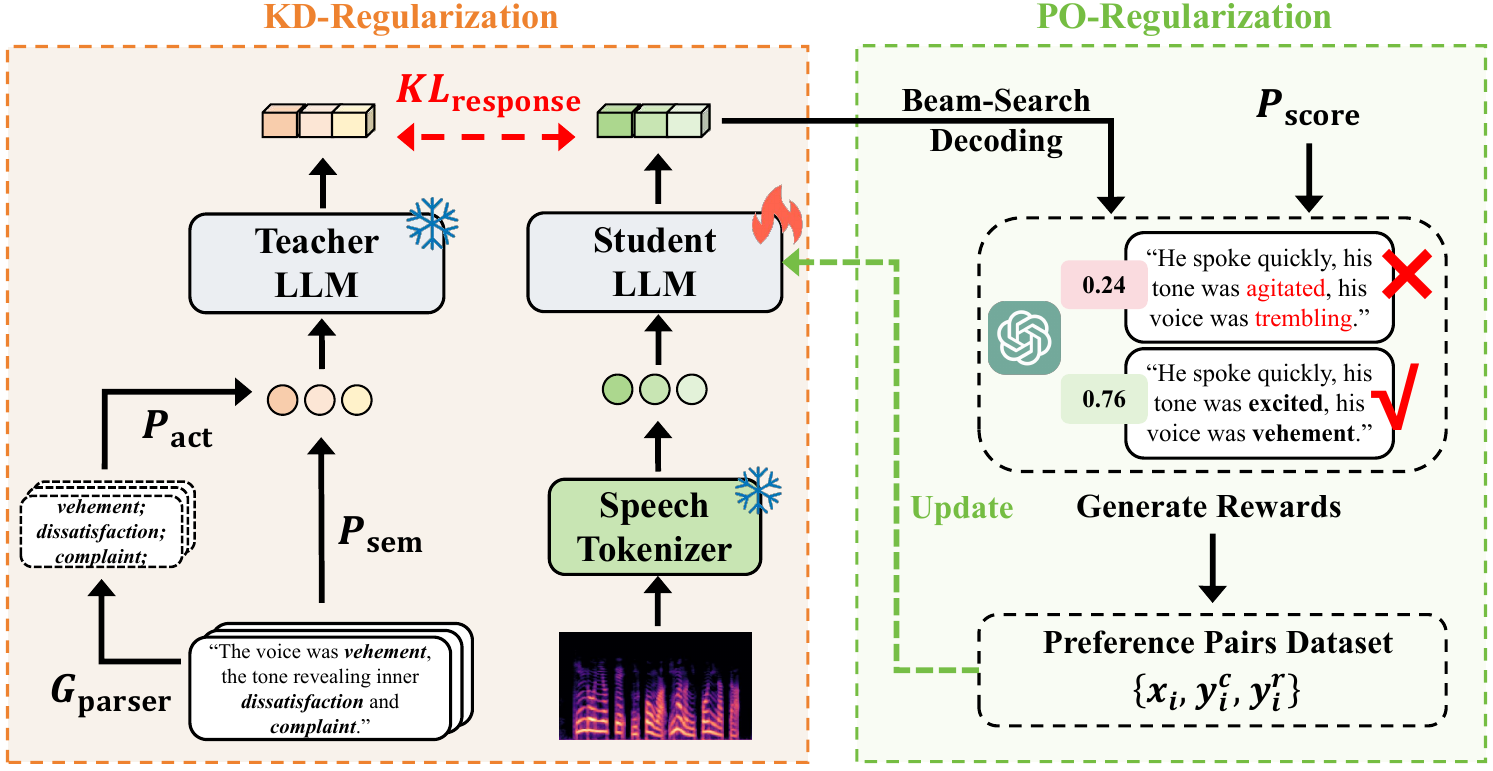}
    \caption{The framework of AlignCap. \textbf{Left:} Illustration of Knowledge Distillation Regularization. Acoustic prompt \textbf{P}$_{\mathrm{act}}$ is generated from emotional clues, which is extracted by an emotion grammar parser \textbf{G}$_{\mathrm{parser}}$. Semantic prompt \textbf{P}$_{\mathrm{sem}}$ is generated from LLM tokenizer. \textbf{Right:} Illustration of Preference Optimization Regularization.} 
    \label{framework}
\end{figure*}

To address these problems, we propose KD-Regularization which achieve Speech-Text Alignment and bridge the distribution gap after LLM decoding. It use the KL-divergence of next-token prediction distributions between LLM's response as a measure of Speech-Text Alignment. As shown in Fig~\ref{distribution}, we observe that align after LLM decoding using knowledge distillation can more effectively improve the speech-text alignment performance.

\section{AlignCap}
\subsection{KD-Regularization}
\label{sec:section3.1}

Our goal is to generate speech emotion captions for speech clips, we design a student LLM to implement speech tokens to text generation and employ a teacher LLM's response to guide student LLM's next-token generation. LLaMA-7B \cite{llama1} is chosen to implement this decoding process due to its exceptional language understanding and modeling capabilities. We simply choose rank value of 8 for LoRA fne-tuning~\cite{hu2022lora} conducted on Student-LLaMA, while the Teacher-LLaMA parameters are frozen.

\para{Acoustic Prompt.}
We first construct a vocabulary of emotional clues, adjectives such as the speaker's tone, intonation, pitch, rhythm, and volume in captions are all regraded as emotional clues. We design an emotion grammar parser (based on NLTK toolkit) to recognize these clues, which are filtered by the vocabulary. Then these clues are inserted into a prompt template P$_{\mathrm{T}}$: <Feeling $e_{1},e_{2},...,$ and $ e_{n}$>, where $e_{n}$ is the $n_{th}$ emotion entity. The acoustic prompt can capture rich and delicate emotion information in emotional clues. 
It can enrich fine-grained emotional description and enhance the robustness of zero-shot captioning for unseen speech, leveraging its training-agnostic nature, which is denoted as:
\begin{equation}
    \begin{split}
        &e_{1\sim n} = G_{\mathrm{Parser}}(y_{i}=\{ c_{i}^{1},...,c_{i}^{|y_{i}|}\})     \\
        &\mathrm{P_{act}} = \mathrm{Insert}(\mathrm{P_{T}}, idx, e_{1\sim n}) \\
    \end{split}
\end{equation}
Where $y_{i}$ is a series of captions, $c_{i}^{m}$ is the $m_{th}$ caption of $y_{i}$. $G_{\mathrm{Parser}}$ and P$_{\mathrm{T}}$ represent emotion grammar parser and prompt template respectively. We insert the emotional clues $e_{1\sim n}$ into the index position $idx$ of P$_{\mathrm{T}}$ to get acoustic prompt P$_{\mathrm{act}}$.


\para{Text Token Generation.}
We denote the captions in speech-caption pairs as the semantic prompt P$_{\mathrm{sem}}$ and concat P$_{\mathrm{act}}$ and P$_{\mathrm{sem}}$ as a prefix prompt, then we provide the prefix prompt along with an instruct prompt (user's instructions) to the LLM to condition its subsequent generation of speech emotion captions using prefix language modeling. 
This setup leverages external knowledge and the language understanding and modeling capabilities of the teacher-LLM to guide the student-LLM to generate plausible sentences.

Given a caption $c_{i}$ with token $T_i$, language model $P_{\theta}$ learns to reconstruct $y_i$ conditioned on the P$_{\mathrm{act}}$ and P$_{\mathrm{sem}}$. The probability of generating the next token is calculated as follows:
\begin{equation}\label{lm_model}
    p_{\theta}(T_t \mid
    \underbrace{T_{0\sim a-1}}_{\mathrm{P_{act}}},
    \underbrace{T_{a\sim b-1}}_{\mathrm{P_{sem}}}, 
    \underbrace{T_{b\sim c-1}}_{\mathrm{P_{instruct}}},
    \underbrace{T_{c\sim t-1}}_{\text{autoregressive}})
\end{equation}
This process is iterated until the LLM generates a token containing a period and the training loss is the maximum likelihood estimate, and the next token $T_t$ is selected according to:
\begin{equation}
\label{texttoken}
    \begin{split}
        T_t=\mathop{\arg\max}_{i \in 1, \ldots, k}
        \Big\{
        p_{\theta} \left(c_i \mid p_n, T_0,\!\ldots\!,T_{t-1} \right) 
        \Big\}
    \end{split}
\end{equation}
Where prefix prompt $p_n$ = P$_{\mathrm{act}}$ $\oplus$ P$_{\mathrm{sem}}$ $\oplus$ P$_{\mathrm{instruct}}$.
Trained on limited data, simply using semantic prompt as prefix prompt may overfit the In-Domain dataset, leading to significant domain shift and performance degradation of language model using out-of-domain (OOD) speech. 
In contrast, the acoustic prompt based on emotion-aware clues, inherits the powerful transferability from captions.

\para{Speech Token Generation.} 
For each speech, we adopt the pre-trained SpeechTokenizer~\cite{zhang2024speechtokenizer} to extract discrete representations and denote the tokens of the first  residual vector quantization (RVQ) layer as speech tokens. 
The first layer of RVQ can be regarded as a semantic token, which contains more content information from speech, resulting in capturing semantically accurate emotional clues. We append this speech token $x_t$ to LLM's input and generate the next token in an autoregressive modeling manner, for each time step $t$, the next token $T_{t}$ is selected according to:
\begin{equation}
\label{speechtoken}
    \begin{split}
        T_t=\mathop{\arg\max}_{i \in 1, \ldots, k}
        \Big\{
        p_{\theta} \left(c_i \mid x_t, T_0,\!\ldots\!,T_{t-1} \right) 
        \Big\}
    \end{split}
\end{equation}

\para{Modality Alignment.} 
Modality adapters~\cite{zeroshotaudiocap2,wavllm} are often used to compress the speech encoder's feature representations. Similar to \cite{multitokenizer}, we treat the input from speech and text modality as a token sequence and learn a joint embedding space for all modalities.
Speech tokens are expanded to text token's codebook in advance so that text and speech share the same codebook.
We pad the shorter token sequence to make it the same length as the longer token sequence. We use a mask to ignore the padding part, ensuring that the model only focuses on valid tokens.



\para{Knowledge Distillation.}
As shown in Fig~\ref{framework}, given a $\mathcal{D}_{s}\!=\!\{(x_{n},y_{n})\}$, we treat the LLM's prediction distribution $p_{\theta} \left(y_n \!\mid\!p_n, y_{<n} \right)$ of the next response token, after having observed the text input $p_n$ and generated partial response $\{y_0,\!\ldots,\!y_{n-1}\}$, as the teacher distribution. Where $p_n$ is the concatenation of P$_{\mathrm{act}}$ and P$_{\mathrm{sem}}$.
In contrast, we consider the corresponding distribution $p_{\theta} \left(y_n \!\mid\!x_n, y_{<n} \right)$ for the speech input $x_n$ as the student distribution.
If speech and text are well-aligned, the two distributions should be close to each other, as measured by KL-divergence, which is as followed:
\begin{equation}
    \begin{split}
        &\mathop{\min}_{\mathrm{LLM_{stu}}(\cdot)}\mathcal{L}_{\mathrm{KL}}(p,x,y)= \\ 
        &-\sum_{t,y_n}p_{\theta}\left(y_n|p_n,y_{<n}\right)\log p_{\theta}\left(y_n|x_n,y_{<n}\right)
    \end{split}
\end{equation}

$\mathcal{L}_{\mathrm{KL}}$ introduces a quantitative measure of speech-text alignment at each step of the response generation process. By minimizing this loss, we can learning a student LLM using LoRA fine-tuning \cite{hu2022lora} for speech input that facilitates generation behaviors similar to those of text inputs when generating speech emotion captions.

\subsection{PO-Regularization}
High-quality emotional description needs to consider not only the richness of emotions but also aspects such as consistency and rationality. The alignment of SEC's output to human preferences is often neglected.
There is a problem that the LLM's response is inconsistent with the user's instructions (faithfulness hallucination) and results in false emotional descriptions (factuality hallucination). Therefore, we propose PO-Regularization to solve these problems.

\para{Preference Pairs Creation.}
Inspired by \cite{instructgpt,self_reward}, we construct a preference pairs dataset by utilizing GPT-3.5 scoring prompt P$_{score}$ on LLM's beam-search decoding output. The P$_{score}$ to act as reward model is used to create preference pairs, which is as follow:

\begin{figure}[h]
    \centering
    \includegraphics[width=7.5cm]{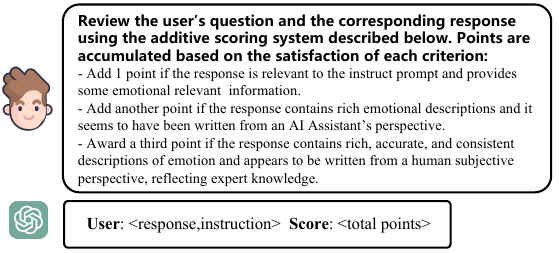}
    \caption{Scoring prompt for candidate responses.} 
    \label{gpt}
\end{figure}

Following above steps, we can get the preference pairs dataset $D_{p}=\{(x_n,y_{n}^{c},y_{n}^{r})\}_{n=1}^{N}$, which is consisted of chosen response $y_{n}^{c}$ and rejected response $y_{n}^{r}$. Finally, we select the highest score as the $y_{n}^{c}$ and the rest as $y_{n}^{r}$.


\para{Preference Optimization.}
To solve the hallucination problem of LLMs, the prevalent RLHF methods \cite{instructgpt,llama2,UltraFeedback} involve fitting a reward model on the preference data, and the training the policy, value and critic models to maximize the reward without deviating too far from the reference model. However, RLHF method contains four models and has too many hyperparameters, making the training complex and high computation cost. Inspired by DPO \cite{DPO,self_reward}, We propose a simpler equivalent supervised approach PO-Regularization that addresses this reinforcement learning goal, the policy model can be directly optimized on the reward feedback based on preference pairs:
\begin{equation}
\label{dpo}
\begin{split}
    \mathcal{L}_{\mathrm{PO}}\!=\!\mathbb{E}_{(x,y_g,y_{n}^{c})}
    \Big[&\beta\log\sigma(\log\frac{\pi_{\theta}(y_g|x)}{\pi_{ref}(y_g|x)} \\ 
    &- \log\frac{\pi_{\theta}(y_{n}^{c}|x)}{\pi_{ref}(y_{n}^{c}|x)})  \Big]
\end{split}
\end{equation}
Where $\beta$ is is a hyperparameter and we only update the policy model $\pi_{\theta}(y|x)$ during finetuning, while reference model $\pi_{ref}(y|x)$ is the same as $\pi_{\theta}(y|x)$ which is frozen to prevent over-optimizing. PO-Regularization considers the likelihood of the preferred response $y_{n}^{c}$ over dispreferred response $y_{n}^{r}$ and optimizes the LLM towards this objective.

\section{Experiments}
\subsection{Dataset}
We select speech-caption paired samples from the large-scale video emotion reason dataset MER2023 \cite{MER2023} to form the MER23SEC dataset. A Chinese interactive multimodal emotion corpus NNIME \cite{NNIME} is used to evaluate the transferability of our model trained on other datasets.
Due to the lack of publicly available high-quality SEC task datasets, we propose a new dataset named EMOSEC\footnote{The EMOSEC dataset is accessible through: \href{https://zenodo.org/records/10948423}{https://zenodo.org/records/10948423}}, which is about 41 hours of Chinese-English Speech Emotion Captioning datasets. It consists of 15 male and 15 female speakers and covers 45039 sentences, with a sampling rate of 16kHz.
We divide MER23SEC, EMOSEC, and NNIME datasets into training, validation and testing according to the ratio of 8:1:1.

\begin{table*}[t]
  \centering
  \resizebox{1.0\textwidth}{!}
  {
  \begin{tabular}{ccccccccc}
    \toprule[1.5pt]
    \multirow{2}{*}{\textbf{\large Dataset}} & \multirow{2}{*}{\textbf{\large Methods}} &\multirow{2}{*}{\textbf{BLEU$@$4}$\uparrow$} &\multirow{2}{*}{\textbf{METEOR}$\uparrow$} &\multirow{2}{*}{\textbf{ROUGE}$\uparrow$} &\multirow{2}{*}{\textbf{CIDEr}$\uparrow$} &\multirow{2}{*}{\textbf{SPICE}$\uparrow$}&\multirow{2}{*}{\textbf{AES$_{c}$}$\uparrow$} &\multirow{2}{*}{\textbf{AES$_{s}$}$\uparrow$}  \\
      &    & & & &   & & &   \\
    \midrule
    \multirow{6}{*}{NNIME}
    & HTSAT-BART~\cite{HTSAT-BART}     & 3.2$\pm$0.5 &8.6$\pm$0.3 &15.7$\pm$0.4 & 2.7$\pm$0.2 & 3.0$\pm$0.4 & 2.5$\pm$0.4 & 3.6$\pm$0.3    \\  
    & NoAudioCap~\cite{zeroshotaudiocap2}   & 4.9$\pm$0.4 &10.4$\pm$0.2 &17.6$\pm$0.3 & 4.9$\pm$0.5 & 5.1$\pm$0.3 & 3.6$\pm$0.2 & 4.2$\pm$0.1    \\
    & SECap~\cite{SECap}  &5.8$\pm$0.4 &11.4$\pm$0.3 &17.9$\pm$0.2 &8.6$\pm$0.4 &5.3$\pm$0.3 &4.9$\pm$0.5 &4.5$\pm$0.2  \\
    \cmidrule(l){2-9} 
    & SECap-PO  &6.0$\pm$0.3 &12.1$\pm$0.3 & 18.6$\pm$0.3 & 8.9$\pm$0.2 & 5.4$\pm$0.4 &5.1$\pm$0.1 & 4.7$\pm$0.4    \\
    & AlignCap-KD-RLHF  &6.6$\pm$0.5 &14.6$\pm$0.5 & 20.9$\pm$0.2 & 9.3$\pm$0.4 & 5.6$\pm$0.2 &5.8$\pm$0.3 & 5.0$\pm$0.2    \\
    & AlignCap-KD-PO  &\bfseries7.7$\pm$0.3 &\bfseries17.3$\pm$0.4 &\bfseries24.3$\pm$0.4 &\bfseries12.8$\pm$0.5 &\bfseries6.4$\pm$0.3 &\bfseries7.3$\pm$0.3 &\bfseries5.6$\pm$0.4   \\
    \midrule
    \multirow{6}{*}{EMOSEC}
    & HTSAT-BART~\cite{HTSAT-BART}   & 4.5$\pm$0.3 &11.6$\pm$0.2 &20.4$\pm$0.5 & 5.1$\pm$0.4 & 3.7$\pm$0.5 & 3.6$\pm$0.2 & 4.8$\pm$0.3  \\  
    & NoAudioCap~\cite{zeroshotaudiocap2}  & 6.7$\pm$0.3 &14.5$\pm$0.4 &21.8$\pm$0.6 & 10.3$\pm$0.6 & 5.7$\pm$0.3 & 4.7$\pm$0.3 & 5.4$\pm$0.1  \\
    & SECap~\cite{SECap}  &7.4$\pm$0.3 &16.6$\pm$0.2 &25.9$\pm$0.3 &11.2$\pm$0.3 &5.8$\pm$0.3 &5.9$\pm$0.3 &5.6$\pm$0.2  \\
    \cmidrule(l){2-9}  
    & SECap-PO  &7.5$\pm$0.4 &17.0$\pm$0.4 & 26.2$\pm$0.2 & 11.8$\pm$0.2 & 6.0$\pm$0.3 &6.1$\pm$0.4 & 5.8$\pm$0.2  \\
    & AlignCap-KD-RLHF  &7.8$\pm$0.3 &18.3$\pm$0.1 & 27.9$\pm$0.4 & 14.2$\pm$0.4 & 6.3$\pm$0.5 &7.0$\pm$0.2 & 6.1$\pm$0.3  \\
    & AlignCap-KD-PO &\bfseries9.8$\pm$0.2 &\bfseries20.9$\pm$0.3 &\bfseries29.8$\pm$0.5 &\bfseries18.7$\pm$0.3 &\bfseries7.6$\pm$0.4 &\bfseries8.8$\pm$0.2 &\bfseries7.6$\pm$0.1  \\
    \bottomrule    
    \cmidrule(r){1-9}
  \end{tabular}
  }
  \caption{Zero-shot evaluation result of different SEC methods on NNIME and EMOSEC.}
  \label{zeroshotSEC}
\end{table*}

\subsection{Settings}
\begin{figure}[t]
    \centering
    \includegraphics[width=7.4cm]{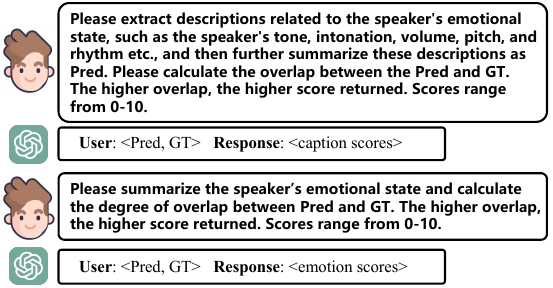}
    \caption{Prompt for Automatic Evaluation.} 
    \label{autoeval}
\end{figure}

\para{Evaluation Metrics.}
We use GPT-3.5 to evaluate the degree of overlap of emotional clues and summarized states as shown in Fig~\ref{autoeval}. The automatic evaluation indicators are denoted as AES$_{c}$ and AES$_{s}$ respectively. The higher the score, the higher quality of generated captions.

To evaluate the accuracy of the generated caption, we initially adopt traditional supervised metrics for the Automated audio captioning (AAC) task, containing standard natural language generation metrics BLEU(B@4), METEOR(M), ROUGE-L(R), CIDEr(C)~\cite{CIDEr}, and SPICE(S)~\cite{SPIDEr}. 
B@4 focuses on the appearance frequency of emotional clues and is used to evaluate the emotional consistency and fine-grainedness of generated captions. 
Compared with B@4, M considers synonyms more, and R pays more attention to the sufficiency and faithfulness of output. C and S Compute accuracy of emotion captions using human consensus.
Therefore, M can be used to evaluate factuality hallucinations, while R, C, and S is used to evaluate faithfulness hallucinations.

\para{Baseline Systems.}
We compare our model with other systems. 1) HTSAT-BART \cite{HTSAT-BART}: a three-stage processing framework, which performs exceptionally well in the AAC task. 
2) NoAudioCap \cite{zeroshotaudiocap2}: a weakly-supervised audio captioning model which requires a pre-trained CLAP \cite{clap}.
3) SECap \cite{SECap}: the fisrt SEC model to generate high-quality speech emotion captions.

\para{Training.}
For KD-Regularization, we optimize the student-LLM with the AdamW optimizer and the learning rate of 1e-5 on 4*V100 GPUs over 50k iterations, the batch size is 16. We employ DeepSpeed \cite{ZeRO} and LoRA \cite{lora} of rank 8 to implement model parallelism and parameter equivalence, applying warmup with 400 steps and gradient accumulation with 8 steps. For PO-Regularization, the learning rate is set to 5e-7 and train for 1000 steps.

\subsection{Main Results}
For Zero-shot scenario, we conduct our model with baselines on NNIME \cite{NNIME} and EMOSEC dataset. Moreover, we evaluated the effects of two different Human preference alignments RLHF-PPO and DPO, on eliminating the hallucinations.

\para{Quantitative Evaluation.}
The objective and automatic evaluation about zero-shot SEC methods are shown in Table~\ref{zeroshotSEC}, and we randomly select 25 sentences from test set to calculate scores. Our proposed preference-optimized models, AlignCap-KD-RLHF and AlignCap-KD-PO, outperform the baseline model in all metrics.
The B@4 and M of AlignCap-KD-PO is higher than that of SECap-PO, which suggests that KD-Regularization can enhance the accuracy of emotional clues modeling. 
The highest R, C, and S scores demonstrate that AlignCap's output exhibits greater sufficiency and faithfulness compared to other baselines. The metrics of SECap-PO is higher than that of SECap, it is attributed to the PO-Regularization, which eliminates the faithfulness hallucinations where the output is inconsistent with user instructions.
AlignCap-KD-PO achieves the highest B$@4$ score, demonstrating that emotional clues as P$_{\mathrm{act}}$ can generate more fine-grained emotion captions. It outperforms AlignCap-KD-RLHF, indicating superior performance in quantitative evaluation. This confirms that DPO-based PO-Regularization can enhance the quality of the caption generated by the model than RLHF-PPO. It also demonstrates that human preference alignment is an effective method for the SEC model to undergo self-improvement.

Compared with NoAudioCap \cite{zeroshotaudiocap2} which also utilizes a similar text-only training method, both AlignCap-KD-RLHF and AlignCap-KD-PO comprehensively surpass NoAudioCap, attributed to our proposed KD-Regularization in alleviating speech-text distribution gap after LLM decoding. Zero-shot inference used by NoAudioCap has a training-inference mismatch, which loses generalization on unseen speech. The KL-divergence in KD-Regularization is used to bridge the mismatch, it suggests that AlignCap can be generalized to unseen speech.


\begin{figure*}[t]
    \centering
    \includegraphics[width=16.2cm]{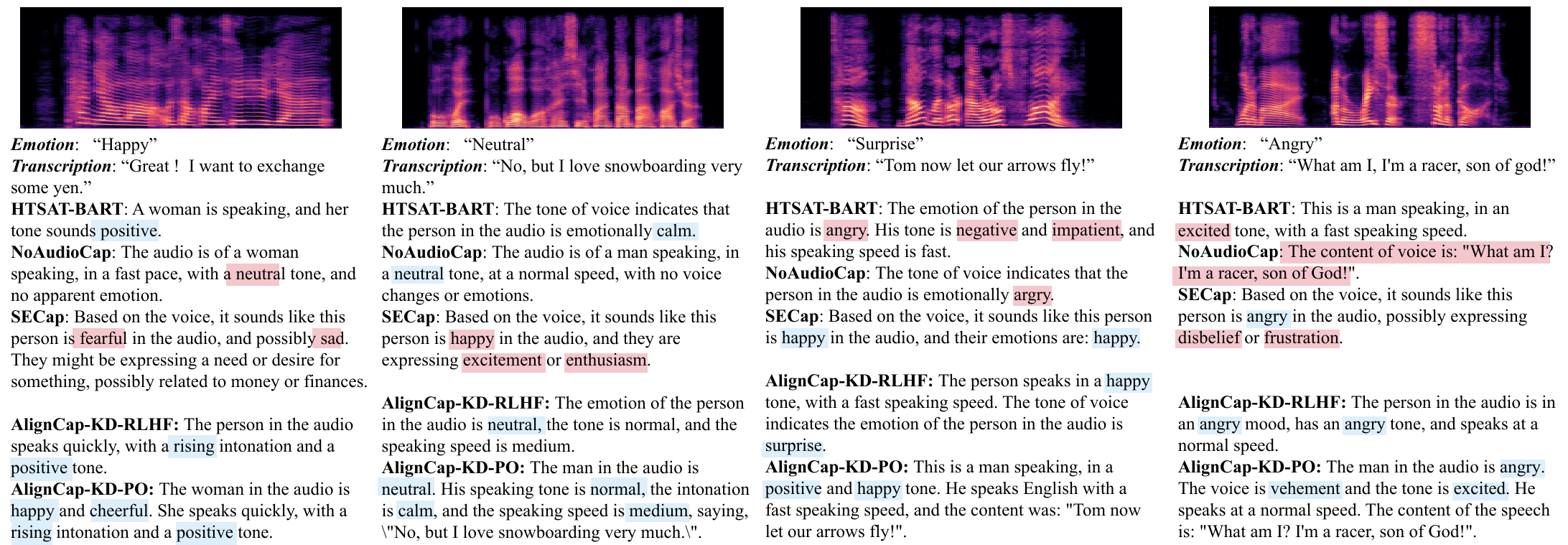}
    \caption{Qualitative Results of Zero-shot SEC with different methods. Incorrect emotional clues in captions are highlighted in \colorbox{backred}{red}, while correct emotional clues in captions are in \colorbox{backblue}{blue}.} 
    \label{demo}
\end{figure*}

\para{Qualitative Evaluation.}
Figure \ref{demo} supports the findings of Table~\ref{zeroshotSEC} by presenting the output of AlignCap and HTSAT-BART \cite{HTSAT-BART}, NoAudioCap \cite{zeroshotaudiocap2}, SECap \cite{SECap}.
Our method can produce richer emotional clues and more coherent emotion captions.
In the “Neutral” example of Figure~\ref{demo}, although SECap~\cite{SECap} can produce rich speech emotion captions, its incorrect emotional cues are inconsistent with the real emotion.
In the “Surprise” example, the output of NoAudioCap lacks fine-grained captions of the speaker's gender, tone, and intonation. AlignCap-KD-PO not only makes up for this shortcoming but also outputs the speaker's content consistent with the transcribed text, which enhances the understanding of the speech content.

In the “Angry” example in Figure~\ref{demo}, AlignCap-KD-RLHF simply refers to gender as "a person", AlignCap-KD-PO can correctly identify its gender by adopting preference optimization, it is also attributed to P$_{\mathrm{act}}$ for enriching fine-grained information about the speaker.

What's more, NoAudioCap~\cite{zeroshotaudiocap2} suffers from LLM's output inconsistent with user instructions, and both AlignCap-KD-PO and AlignCap-KD-RLHF eliminate this faithfulness hallucination, owing to our proposed PO-Regularization.

\subsection{Ablation Studies}

As shown in Table~\ref{Ablation}, we train AlignCap with specific components selectively removed to evaluate the effect of the proposed components to eliminating hallucinations and enrich fine-grained information.

The decrease in all objective evaluation scores shows the significance of acoustic prompt (P$_{\mathrm{act}}$), KD-Regularization ($\mathcal{L}_{\mathrm{KL}}$), and PO-Regularization ($\mathcal{L}_{\mathrm{PO}}$). 
The significant decrease of P$_{\mathrm{act}}$ on B@4 proves the positive effect of the emotional clues extracted by P$_{\mathrm{act}}$ on the emotional consistency of generated captions. 
The lack of $\mathcal{L}_{\mathrm{KL}}$ and $\mathcal{L}_{\mathrm{PO}}$ leads to a significant drop in M and R, indicating that they play a crucial role in guiding Zero-shot SEC model to eliminate factuality and faithfulness hallucinations.
The model without P$_{\mathrm{act}}$ also exhibits a decrease in R, indicating its effectiveness in generating fine-grained emotional descriptions.
The M score of AlignCap-KD-PO is higher than AlignCap-KD-RLHF without adopting explicit reward modeling, allowing it to learn more human-like generated captions.

\begin{table}[t!]
\centering
\resizebox{0.48\textwidth}{!}{
    \begin{tabular}{c|ccc|cccc}
        \toprule[1.5pt]
         \multirow{2}{*}{\textbf{Methods}}&\multirow{2}{*}{\textbf{P$_{\mathrm{act}}$}} &\multirow{2}{*}{\textbf{$\mathcal{L}_{\mathrm{KL}}$}} & \multirow{2}{*}{\textbf{$\mathcal{L}_{\mathrm{PO}}$}}& \multicolumn{4}{c}{\textbf{EMOSEC}}\\
         &&& &B@4$\uparrow$ &M$\uparrow$ & R$\uparrow$ & C$\uparrow$  \\
         \midrule
        \multirow{4}{*}{AlignCap-KD-RLHF}
        &\ding{52} &\ding{52} &\ding{52} &/    &/    &/    &/     \\
        &-         &\ding{52} &\ding{52} &-3.5 &-2.3 &-2.7 &-1.2  \\
        &\ding{52} &-         &\ding{52} &-1.9 &-5.4 &-4.3 &-1.9   \\
        &\ding{52} &\ding{52} &-         &-1.5 &-2.7 &-3.2 &-1.6     \\
        \midrule
        \multirow{4}{*}{AlignCap-KD-PO}
        &\ding{52} &\ding{52} &\ding{52} &/    &/    &/    &/     \\
        &-         &\ding{52} &\ding{52} &-1.4 &-1.3 &-1.8 &-0.7  \\
        &\ding{52} &-         &\ding{52} &-0.8 &-4.7 &-3.0 &-1.6  \\
        &\ding{52} &\ding{52} &-         &-0.5 &-2.6 &-2.2 &-0.9    \\
        \bottomrule[1.5pt]
    \end{tabular}
    }
    \caption{Ablation studies on EMOSEC dataset.}
\label{Ablation}
\end{table}

\section{Analysis}

\subsection{Transferability on Cross-Domain Speech}

As shown in Table~\ref{crossdoamin}, we evaluate the AlignCap in a cross-domain scenarios where the training data and testing data are from different datasets.
We conduct experiments on NNIME and MER23SEC's testing set, and we only use the training set of EMOSEC's captions to fine-tuning AlignCap.

\begin{table}[t!]
\centering
\resizebox{0.49\textwidth}{!}{
    \begin{tabular}{cccc}
        \toprule[1.5pt]
        \multirow{1}{*}{\textbf{Methods}} &\multirow{1}{*}{\textbf{B@4$\uparrow$}} &\multirow{1}{*}{\textbf{M$\uparrow$}} &\multirow{1}{*}{\textbf{R$\uparrow$}}\\ 
        \midrule
        &      & \textbf{EMOSEC$\rightarrow$NNIME} &   \\
        \cline{2-4} 
        HTSAT-BART &1.9$\pm$0.4 &3.4$\pm$0.5 &6.1$\pm$0.3   \\
        NoAudioCap &4.2$\pm$0.2 &8.7$\pm$0.4 &10.3$\pm$0.6  \\
        SECap &3.4$\pm$0.3 &8.2$\pm$0.3 &13.8$\pm$0.5 \\
         AlignCap-KD-RLHF &5.2$\pm$0.4 &9.6$\pm$0.3 &14.4$\pm$0.4     \\
         AlignCap-KD-PO  &\textbf{5.9$\pm$0.3} &\textbf{10.1$\pm$0.5} &\textbf{15.6$\pm$0.2}   \\
         \midrule
         & &\textbf{EMOSEC$\rightarrow$MER23SEC} &\\
         \cline{2-4}
         HTSAT-BART  &4.4$\pm$0.3 &11.1$\pm$0.4 &12.3$\pm$0.4  \\
        NoAudioCap &11.3$\pm$0.3 &13.2$\pm$0.3 &18.7$\pm$0.3  \\
        SECap &9.8$\pm$0.3 &14.6$\pm$0.3 &16.1$\pm$0.3  \\
         AlignCap-KD-RLHF &14.5$\pm$0.2 &19.8$\pm$0.4 &21.0$\pm$0.5   \\
         AlignCap-KD-PO  &\textbf{16.3$\pm$0.3} &\textbf{21.6$\pm$0.3} &\textbf{22.7$\pm$0.3}   \\
        \bottomrule[1pt]
    \end{tabular}
    }
    \caption{Cross-domain SEC results on NNIME and MER23SEC dataset.}
\label{crossdoamin}
\end{table}

In EMOSEC$\rightarrow$NNIME cross-domain scenarios, the results show that AlignCap outperforms all baselines. The B@4 and M metrics of SECap and HTSAT-BART are lower than NoAudioCap on the NNIME dataset. 
This is because they all have encoder-decoder structures and are trained on well-paired data, lacking components to enhance generalization capabilities for cross-domain data. 

AlignCap outperforms NoAudioCap, demonstrating the superiority of the proposed KD-based speech-text alignment over the CLAP-based (contrastive learning) speech-text alignment used in NoAudioCap for cross-modal mapping. It not only bridges the audio-text distribution gap, but also improves the generalization ability in cross-domain scenarios.

Additionally, there is a domain offset between the predicted emotional description generated by LLMs and the real description of the target domain, leading to performance degradation.
Equipped with the PO-Regularization, AlignCap-KD-PO outperforms the other baselines including AlignCap-KD-RLHF version on most metrics, demonstrating the effectiveness of the proposed compoents.

\subsection{Effect of Different Speech-Text Alignment on Downstream SEC task}
\begin{figure}[h]
    \centering
    \includegraphics[width=6.7cm]{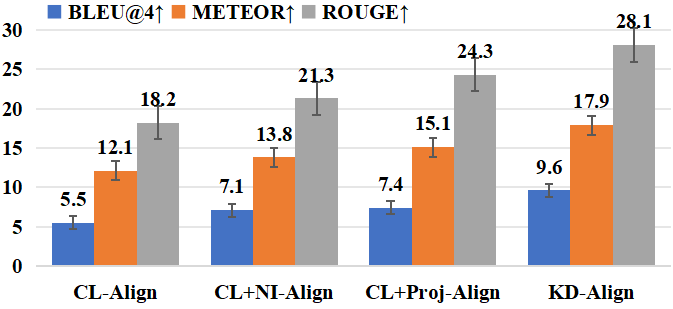}
    \caption{The impact of different Alignment.} 
    \vspace{-0.3cm}
    \label{analysis_align}
\end{figure}

Previous alignment methods, such as Gaussian Noise Injection (CL+NI-Align) \cite{zeroshotaudiocap2} and Project-based Decoding (CL+Proj-Align) \cite{DCASEWorkshop2023}, achieve alignment by adding Gaussian noise variance or mapping based on contrastive learning between speech and text embeddings before LLM decoding. The KD-Regularization (KD-Align) we proposed achieves speech-text alignment after LLM decoding and alleviates the information loss in modality alignment. 
Fig~\ref{analysis_align} shows that our method outperforms other alignment methods in all indicators, attributing to we treat speech-text alignment as a knowledge distillaiton problem. It can ensure that the LLM's responses to speech inputs closely mirror those to corresponding text inputs.

\subsection{Performance on Different Preference Pair Sizes and Steps.}

\begin{figure}[h!]
    \centering
    \begin{minipage}{0.23\textwidth} 
        \includegraphics[width=\linewidth]{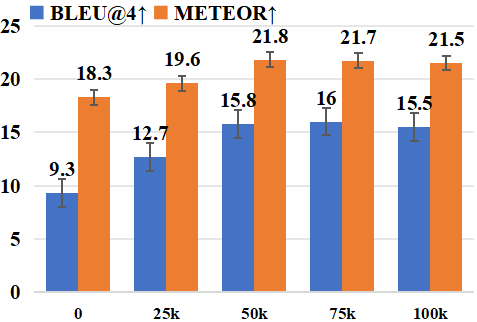}
        \label{fig:analysis3}
    \end{minipage}%
    \hfill 
    \begin{minipage}{0.25\textwidth}
        \includegraphics[width=\linewidth]{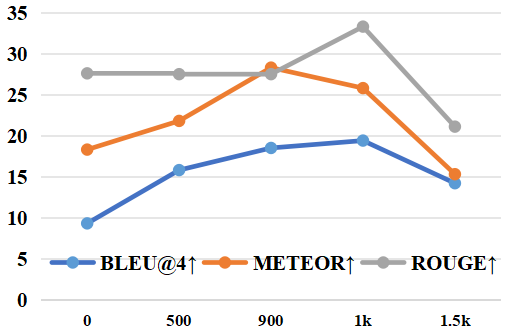}
        \label{fig:analysis3_2}
    \end{minipage}
    \vspace{-0.5cm}
    \caption{\textbf{Left:} Performance of AlignCap across different preference pair sizes. \textbf{Right:} Performance of AlignCap of different fine-tuning steps.}
    \label{analysis_prefer}
    \vspace{-0.3cm}
\end{figure}

As shown in Fig~\ref{analysis_prefer}, we examine the effect of different preference pair sizes and fine-tuning steps for PO-Regularization on the performance of AlignCap. We set the preference pair sizes to be $\{0,25\mathrm{k},50\mathrm{k},75\mathrm{k},100\mathrm{k}\}$. After 500 steps of fine-tuning with DPO for each of these sizes, we assess their performance in zero-shot SEC. We can observe notable improvement with increasing sizes from 0 to 50k, which indicates that an increase in preference pair data can improve zero-shot SEC.
However, using more than 50k preference data for DPO does not lead to significant performance improvements, indicating a threshold beyond which additional data does not enhance learning outcomes.

Moreover, we set the fine-tuning steps to be 0$\rightarrow$1.5k for AlignCap on zero-shot SEC evaluation. As shown in Fig~\ref{analysis_prefer}, all metrics demonstrate significant performance improvement when the number of fine-tuning steps is less than 1k. 
However, when the number of iterations exceeds 1k steps, the model suffers from overfitting, resulting in performance degradation, indicating that 1k steps are the optimal iteration steps for PO-Regularization.

\subsection{Can PO-Regularization Works with Small Models?}

We investigate whether PO-Regularization can bring improvements for smaller language models. The preference pair size is 50k and we fine-tuning the models on EMOSEC dataset for 1k steps. We evaluate the zero-shot SEC performance on EMOSEC test set. Tab~\ref{smallmodel} shows that after 1k iterations, PO-Regularization significantly boosts OPT's scores but decreases the GPT2-base's scores on M and R, while improving GPT-2-large very little.
This indicates that PO-Regularization can improve caption generation in small language models, although the improvement is not significant for models with very small parameters.

\begin{table}[h!]
\centering
\resizebox{0.48\textwidth}{!}{
    \begin{tabular}{ccccc}
        \toprule[1.5pt]
        \multirow{1}{*}{\textbf{Models}} &\multirow{1}{*}{\textbf{Parameters}} &\multirow{1}{*}{\textbf{B@4$\uparrow$}} &\multirow{1}{*}{\textbf{M$\uparrow$}} &\multirow{1}{*}{\textbf{R$\uparrow$}}\\ 
        \midrule
        GPT2-base &\multirow{2}{*}{124M} &3.3$\pm$0.3 &8.4$\pm$0.5 &16.1$\pm$0.3   \\
        GPT2-base-PO        &            &3.3$\pm$0.5 &8.3$\pm$0.4 &16.0$\pm$0.5   \\
        \midrule
        GPT2-large &\multirow{2}{*}{774M}  &3.7$\pm$0.2 &8.9$\pm$0.3 &17.3$\pm$0.4  \\
        GPT2-large-PO &  &3.8$\pm$0.3 &9.1$\pm$0.4 &17.6$\pm$0.2  \\
        \midrule
        OPT       &\multirow{2}{*}{1.3B}  &4.3$\pm$0.2 &10.2$\pm$0.4 &18.8$\pm$0.5 \\
        OPT-PO       &  &5.4$\pm$0.3 &12.8$\pm$0.2 &20.5$\pm$0.4 \\
        \bottomrule[1pt]
    \end{tabular}
    }
    \vspace{-0.1cm}
    \caption{Performance on Small Models.}
\vspace{-0.3cm}
\label{smallmodel}
\end{table}

\section{Conclusion}
We proposed AlignCap, achieving speech-text alignment and human preference alignment. To minimize the distribution gap between LLM's response to speech input and those to corresponding text inputs, we design KD-Regularization to achieve speech-text alignment.
Additionally, we align emotion captions to human preference by PO-Regularization. This process eliminates the factuality and faithfulness hallucinations of AlignCap on unseen speech.
Experiments demonstrate AlignCap’s superiority in both zero-shot and cross-domain scenarios.

\section*{Limitations}
Well-paired speech-caption datasets are difficult to obtain in real-world scenarios. Captions containing emotional descriptions are easy to obtain, but high-quality speech-caption paired data is difficult to collect, how to solve this mismatch problem will be left to our future work. In addition, enhancing the robustness of alignment between speech and text inputs remains an urgent issue that needs to be addressed in the future.

\bibliography{emnlp2024}
\bibliographystyle{acl_natbib}

\end{document}